\pdfoutput=1

\documentclass[11pt]{article}

\usepackage{ACL2023}

\usepackage{times}
\usepackage{latexsym}
 \usepackage{mathrsfs}
 \usepackage{graphicx}
  \usepackage{booktabs}
 \usepackage{color}
 \usepackage{multicol}
\usepackage{multirow}
\usepackage[T1]{fontenc}
\usepackage{diagbox}

\usepackage[utf8]{inputenc}

\usepackage{microtype}

\usepackage{inconsolata}

\title{Interactive Molecular Discovery with Natural Language}

\author{Zheni Zeng\textsuperscript{1}, Bangchen Yin\textsuperscript{1}, Shipeng Wang\textsuperscript{2}, Jiarui Liu\textsuperscript{2}, Cheng Yang\textsuperscript{3}, \\
\textbf{Haishen Yao\textsuperscript{2}, Xingzhi Sun\textsuperscript{2}, Maosong Sun\textsuperscript{1}, Guotong Xie\textsuperscript{2*}, Zhiyuan Liu\textsuperscript{1*} } \\
\textsuperscript{1} Tsinghua University \ \ \ \textsuperscript{2} PingAn Technology \\
\textsuperscript{3} Beijing University of Posts and Telecommunications
\\
\textsuperscript{*}\texttt{xieguotong@pingan.com.cn, liuzy@mail.tsinghua.edu.cn}\\
}

\begin{document}
\maketitle
\begin{abstract}
Natural language is expected to be a key medium for various human-machine interactions in the era of large language models. When it comes to the biochemistry field, a series of tasks around molecules (e.g., property prediction, molecule mining, etc.) are of great significance while having a high technical threshold. Bridging the molecule expressions in natural language and chemical language can not only hugely improve the interpretability and reduce the operation difficulty of these tasks, but also fuse the chemical knowledge scattered in complementary materials for a deeper comprehension of molecules. Based on these benefits, we propose the conversational molecular design, a novel task adopting natural language for describing and editing target molecules. To better accomplish this task, we design ChatMol, a knowledgeable and versatile generative pre-trained model, enhanced by injecting experimental property information, molecular spatial knowledge, and the associations between natural and chemical languages into it. Several typical solutions including large language models (e.g., ChatGPT) are evaluated, proving the challenge of conversational molecular design and the effectiveness of our knowledge enhancement method. Case observations and analysis are conducted to provide directions for further exploration of natural-language interaction in molecular discovery.

\end{abstract}

\section{Introduction}
Molecular design is a fundamental task in various fields including biochemistry and material science, and has made great progress in recent years with the development of deep learning technology~\cite{wang2022deep}. Existing systems usually directly generate molecules or optimize the given molecules~\cite{du2022molgensurvey}, in the form of various chemical languages like the simplified molecular-input line-entry system (SMILES)~\cite{weininger1988smiles} and the structural formula. On the one hand, the expression intermediate of chemical languages lacks readability and requires a great deal of human expertise to use. On the other hand, the task form in current molecular design systems lacks interactivity and does not well-integrate operations such as retrieval and editing. Therefore, though with the assistance of deep learning methods, molecular design is still difficult and time-consuming for researchers.

\begin{figure}[t]
    \centering
    \includegraphics[width=0.7\linewidth]{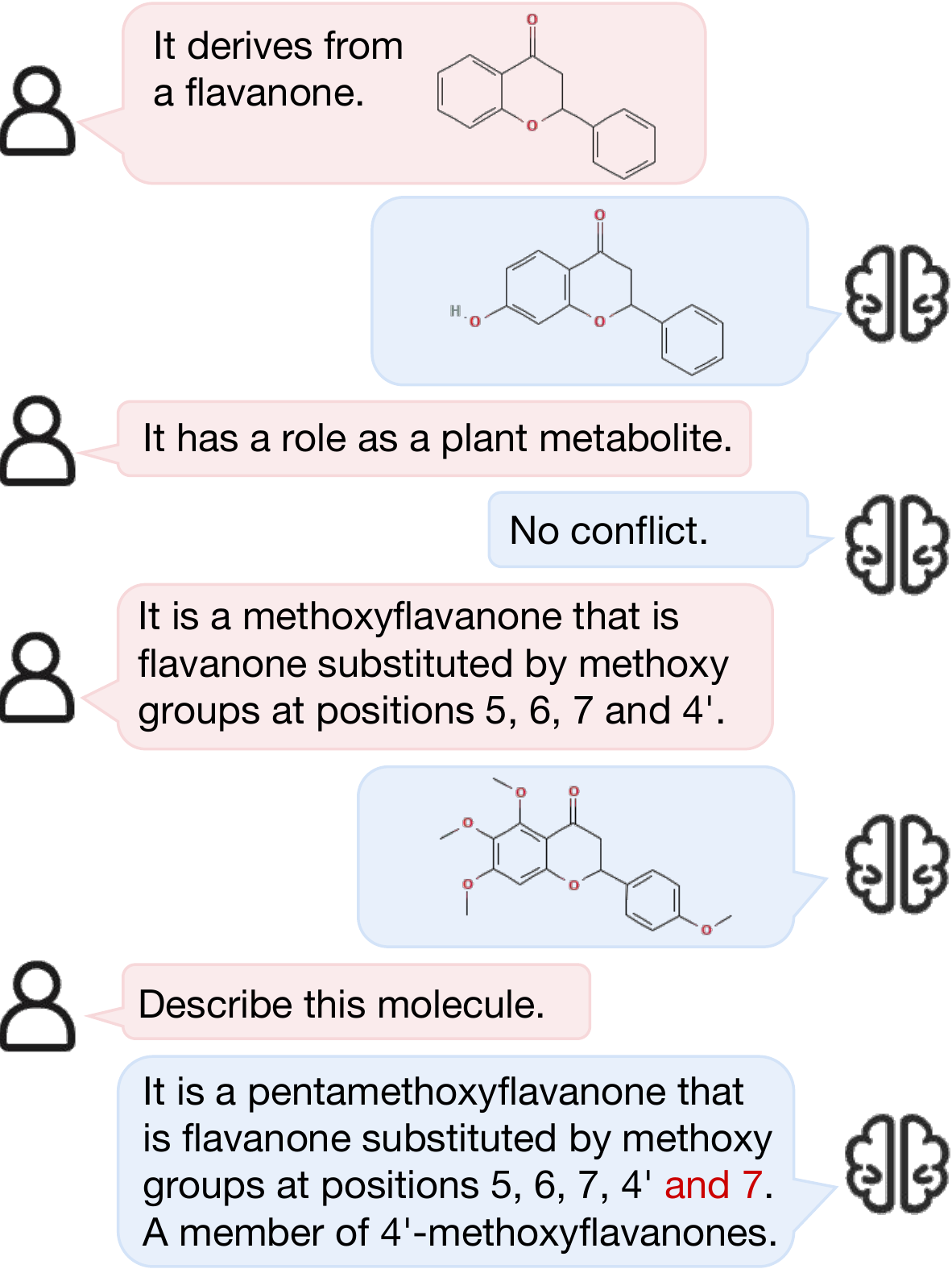}
    \caption{Conversational molecular design diagram.}
    \label{fig:task}
\end{figure}

Luckily, the appearance of pre-training language models (PLMs)~\cite{han2021pre} further brings new possibilities for this field, especially when the large language models (LLMs)~\footnote{\url{https://platform.openai.com/docs/model-index-for-researchers}} come out and show their great capability of providing a mass of biochemical knowledge and executing flexible human instructions~\cite{tong2023discovering}. This inspires us that we may easily communicate with natural language processing (NLP) systems about molecular design requirements. Meanwhile, factual findings about molecules described externally with natural language (e.g., biochemical literature) have to be taken into consideration when understanding molecules, because they can provide complementary information for chemical languages which characterize molecular structures and mechanisms internally. Therefore, it is both worthy and necessary to explore adopting the recent NLP techniques to facilitate molecular design.

Currently, preliminary explorations have been started about transferring NLP methods to molecule tasks, such as the self-supervised architectures for molecule comprehension~\cite{goh2018using, rong2020self, wang2019smiles}. As for using natural language in molecular design, several tasks are proposed including molecule-description matching~\cite{zeng2022deep}, molecule captioning, and generation according to descriptions~\cite{edwardstranslation}. However, these tasks are in various forms while not flexible and scalable enough, and have not maximized the benefit of using natural language as an interaction medium for molecular manipulations. 

We unify existing tasks into an innovative form and propose \textbf{conversational molecular design}, in which the human manipulators freely provide molecular information in chemical or natural language, asking for readable property descriptions or modified molecules that meet the given requirements in multi-turn conversations. To cope with the training and evaluating demand for conversational molecular design, we create the conversation data based on the molecule-description parallel datasets with the help of manual-designed rule filters.  

Conversational molecular design is still a challenging task, though PLMs can be used to understand complicated natural-language interactive commands. To be specific, PLMs are proven to be effective when processing natural and chemical languages isolatedly, while conversational design requires a flexible and synergistic comprehension of two kinds of text. Besides, chemical materials are highly specialized and it is not enough for the models to handle molecular design with only commonsense knowledge.

To achieve the newly proposed task, we design ChatMol, which is a knowledgeable and versatile model based on a generative PLM that can deeply understand and generate natural language~\cite{raffel2020exploring}. To solve the above two challenges, ChatMol is enhanced to first bridge natural and chemical languages smoothly to enable interactive conversation, and then to master plenty of molecular knowledge to better design molecules.

For \textbf{bridging different languages}, we encourage ChatMol to process natural and chemical languages at the same time, capturing complicated associations between two kinds of text with minimum supervision. We first regard SMILES strings as normal texts and conduct masked language modeling on SMILES and biochemical literature materials separately to gain a basic comprehension of both languages. Further, we train the model to read natural language text, recognize chemical entities, and convert to the chemical language expressions of these entities. To automatically construct the training data, we adopt chemical named entity recognition tools to identify entities in the literature and query their SMILES strings from existing records. 

For \textbf{injecting molecular knowledge}, we introduce two types of sophisticated while representative knowledge to assist ChatMol to handle the specialized molecular task: \textit{experimental properties} from knowledge bases (KBs), and \textit{spatial structures} from toolkit calculation. The former includes physical and chemical properties from wet-lab experiments recorded in molecule KBs in the form of natural language. The latter includes the chemical bond, molecular ring and aromaticity information, and this helps the spatial structure information reflected in the SMILES expression become more intuitive. Given the chemical SMILES following the task prefixes, our model is trained to generate the natural language answers constructed with the above information.

Our main contributions are as followed: (1) We propose conversational molecular design, an innovative task adopting natural language as the basic medium of human-machine interaction for conducting low-barrier molecular discovery; (2) We provide ChatMol, which is a pre-trained model that can perform well in the new task with the enhancement of molecular knowledge and chemical-natural language associations; (3) Experiment results prove the challenge of the task we propose and also the effectiveness of our method, achieving better performance with much fewer pre-training steps. 


\section{Conversational Molecular Design}
\label{subsection:mutual}

In this section, we clarify the definition and related settings of the newly proposed task.

\subsection{Task Definition} 
In each turn of the molecular design, both the human user and the intelligent system may provide molecules expressed in chemical language or chemical properties expressed in natural language. Given the conversation history $H(\mathcal{M},\mathcal{T})$ which contains molecules $\mathcal{M}_{1,2,...,p}$  and chemical property descriptions $\mathcal{T}_{1,2,...,q}$, we focus on two main functions that researchers expect the intelligent system to do:

(1) Molecule understanding: The system is required to generate a paragraph of property descriptions $\mathcal{T}'_i$ for the molecule $\mathcal{M}_i$ in $H$. (2) Molecule generation: The system is also supposed to generate a specific molecule $\mathcal{M}'_j$ that meets the requirements in $H$, which may be text descriptions or the molecules that the target has to be similar with. Since there may be more than one molecule matching the descriptions, the human user can iteratively replenish property descriptions $\mathcal{T}_{j+1}$, and the system would generate a modified molecule $\mathcal{M}'_{j+1}$ according to the current text descriptions $\mathcal{T}_{1,2,...,j+1}$ and the last-round result $\mathcal{M}'_j$ (with the input as ``it looks like $\mathcal{M}'_j$'').

\begin{figure*}[ht]
    \centering
    \includegraphics[width=\linewidth]{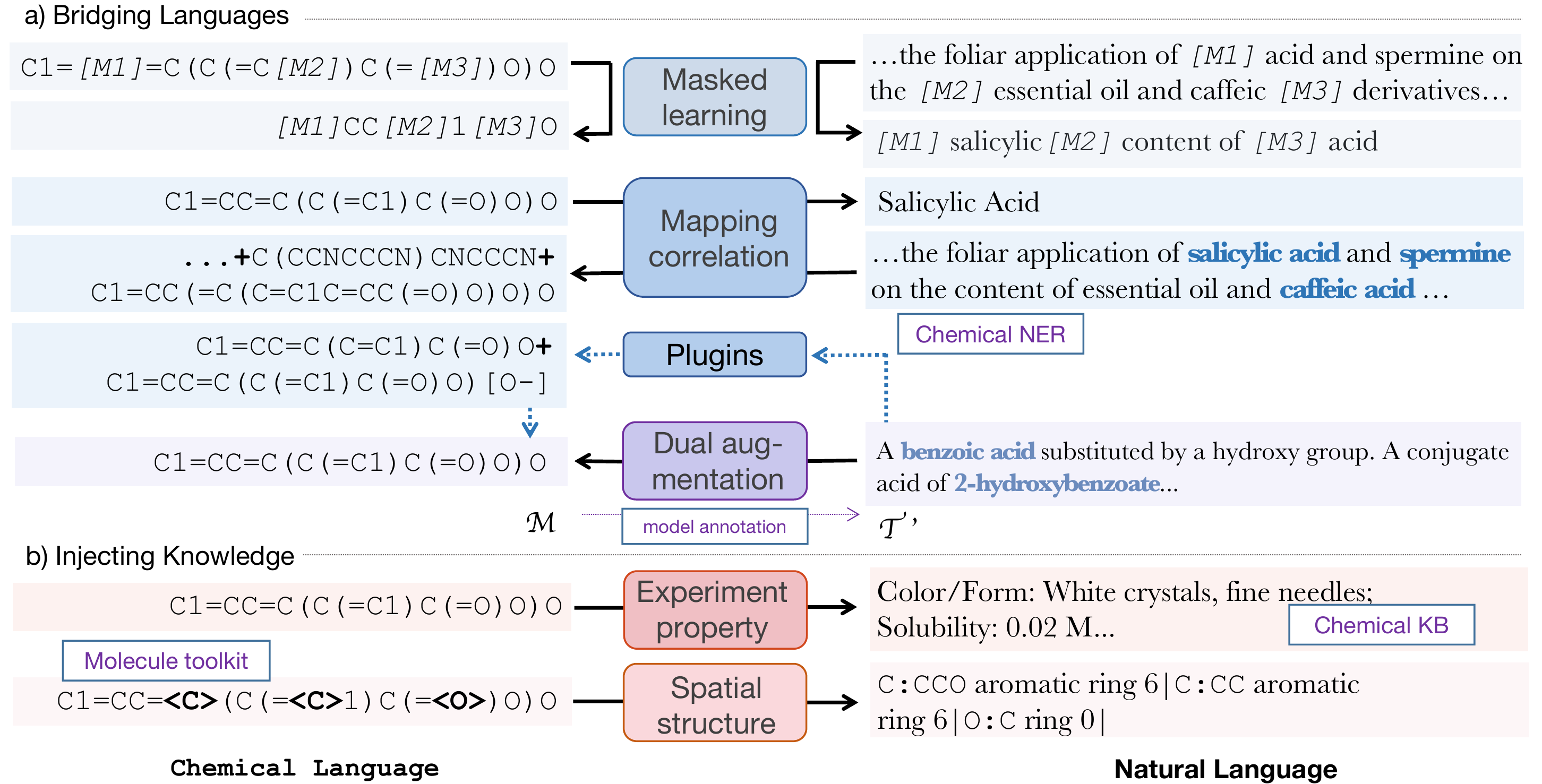}
    \caption{Knowledgeable and versatile training process for ChatMol. The black arrows indicate the input to output direction of the task. }
    \label{fig:model}
\end{figure*}

\subsection{Evaluation Metrics}
\label{subsec:metric}
For traditional generation tasks, evaluation metrics such as BLEU~\cite{papineni2002bleu}, ROUGE~\cite{lin2004rouge} and METEOR~\cite{banerjee2005meteor} are widely used to measure the similarity of the predicted results and the reference answers. They can also be applied to the evaluation of molecule understanding and generation. However, the textual similarity is not persuasive from the perspective of the accuracy of molecules.

For molecule generation, we first judge whether the generated strings are legal molecule SMILES with the help of the RDKit toolkit~\cite{landrum2013rdkit}. 
We pay extra attention to the exact match (hit@1) and hit@3 accuracy. Besides, molecular similarity measuring methods such as various fingerprint Tanimoto similarity (FTS) metrics~\cite{tanimoto1958elementary,schneider2015get,durant2002reoptimization} are also persuasive, since they detect similar functional substructures in the molecules and can indicate the similarity of properties.

\subsection{Model and Data Preparation} 
Since the features of SMILES and natural language are quite different, we prepare two sets of encoder and decoder to separately process $\mathcal{M}$ and $\mathcal{T}$. Each set is set as a common sequence-to-sequence (seq2seq) framework and initialized with T5~\cite{raffel2020exploring}, a widely-adopted seq2seq PLM. 

As for the data preparation, the corresponding ${<}\mathcal{M}, \mathcal{T}{>}$ pairs can be acquired from the chemical KBs, in which brief text introductions of substances are provided. Information leakage may happen due to the appearance of molecule names in the text. To alleviate the problem, synonyms of the target molecule are replaced by general referential phrases such as ``the molecule''. 

To conduct the multi-turn molecule generation, we create a new dataset \textbf{ChEBL-dia} based on ChEBL-20. In the original dataset, $\mathcal{T}$ used to describe the molecule $\mathcal{M}$ usually contains multiple sentences, in the order of description from fine structures to overall properties. To get the multi-turn text descriptions, we split $\mathcal{T}$ into sentences and reverse their order to get $\mathcal{S}_{1,2,...,n}$ in which the amount of detailed information increases sequentially. For the $k_{th}$ turn, we have $\mathcal{T}_k=\{\mathcal{S}_1,\mathcal{S}_2,...,\mathcal{S}_k\}$. To get the molecule intermediate results, we adopt the pre-trained Molt5-caption2smiles-large model~\cite{edwardstranslation} to automatically generate 5 candidates $\mathcal{M}_{k1,k2,...,k5}$ for the given $\mathcal{T}_k$, and randomly pick one as the molecule $\mathcal{M}_{k}$ expected to be generated, which has RDK fingerprint similarity~\cite{landrum2013rdkit} with the final answer $\mathcal{M}_{n}$ higher than 0.5 while lower than 1 (i.e., to avoid information leakage). 

Additionally, we filter out those items that contain only one turn of conversation, and delete those with '-' in the sentences to avoid the appearance of the standard chemical nomenclature (e.g., IUPAC nomenclature) to directly reveal the answer. Besides, we randomly remain a few items that the intermediate molecules have low similarities with the final answers to keep the variety and help enhance the model's robustness.

\section{Methodology}

In this section, we present the knowledgeable and versatile training method of ChatMol which is designed to better finish the conversational molecular design task. The overall process is shown in Figure~\ref{fig:model}. We mix up all the data with different task prefixes to conduct multi-task pre-training, and apply the plugins mentioned in molecule mapping correlation in fine-tuning.

\subsection{Briding Different Languages}
\label{sec:method}

Conducting the popular pre-training paradigm masked language modeling (MLM) ensures the basic comprehension and generation capability isolated on natural and chemical languages. However, to process the multi-modal conversational molecular design data, it is necessary for our model to capture the associations and convert between the two languages. We apply the following two strategies to construct more mapping data for our model.

\textbf{Molecule Mapping Correlation.} We recognize the molecular entities in the literature corpus with the SciSpacy tool~\cite{neumann2019scispacy} and then query for their SMILES from PubChem~\cite{kim2016pubchem}. Given the natural language segments, the molecule generation model is required to generate SMILES expressions for all the molecules that appeared in it sequentially. Conversely, the molecule understanding model is expected to generate the standard names of the given molecules. In this way, we help ChatMol build the parallel association with the least supervision. 

Notice that the toolkit and KB can also play a role as plugins during the downstream inferencing, we provide the automatically-annotated entity SMILES strings as the prompts given text descriptions for molecule generation. To avoid information leakage, we forcefully remove those SMILES strings in the prompts that are the same as molecules in the answers. 

\textbf{Dual Augmentation.} Considering that the molecule understanding (molecule-to-text generation) and molecule-to-text generation are a pair of mutual tasks, while existing molecule SMILES strings are much more than molecular property descriptions in natural language, we can take the idea of dual learning~\cite{he2016dual,xia2017dual}, a common mechanism in neural machine translation, to alleviate the challenge of lacking parallel data. Specifically, the fine-tuned molecule understanding model can generate augmented text descriptions for any given molecule, which can feedback on the molecule generation training.

\subsection{Injecting Molecular Knowledge}

To get an in-depth comprehension of the given chemical language expressions and to generate more reasonable and informative natural language descriptions, our model is injected with two types of molecular knowledge. 

\textbf{Experimental Property Knowledge.} 
We expect the model to understand the molecular structures and to master properties explicitly described in the natural language. We collect 15 types of physical and chemical properties determined experimentally from PubChem database~\cite{kim2016pubchem}, including solubility, color, corrosivity and so on. These properties can be directly described in natural language and provide supervision signals for molecule understanding. 

\textbf{Spatial Structure Knowledge.} Spatial information about molecules is essential for understanding molecular properties. SMILES expressions do not directly express the topological structure of molecules in the language model. To cope with the demand of understanding molecule structures, we introduce spatial-related pre-training tasks. We use the RDKit toolkit~\cite{landrum2013rdkit} to get the spatial structure of the input molecule. For the specific atom, the model is required to recognize the other atoms connected to it, as well as its aromaticity and ring formation information. 

\section{Experiment}
In this section, we first introduce the basic settings including datasets and baseline models. Next, we introduce the molecule understanding and generation experiments and the corresponding ablation study. To evaluate the model capability more comprehensively, we also provide the case analysis.

\begin{table*}[ht]
\centering
\begin{tabular}{l|c|c|cccccc}
\toprule
Model & Training Steps & Dataset& \textbf{BL-2}↑ & \textbf{BL-4}↑ & \textbf{RG-1}↑ & \textbf{RG-2}↑ & \textbf{RG-L}↑ & \textbf{MET}↑  \\
\midrule
\multirow{2}{*}{ChatGPT} & \multirow{2}{*}{-} & ChEBI & 0.226 & 0.121 & 0.311 & 0.117 & 0.232 & 0.292 \\
& & PCdes &  0.144 & 0.055 & 0.244 & 0.062 & 0.158 & 0.233 \\
\hline
\multirow{2}{*}{KV-PLM} & \multirow{2}{*}{0} & ChEBI & 0.406 & 0.319 & 0.506 & 0.323 & 0.430 & 0.506 \\
& & PCdes & 0.235 & 0.145 & 0.331 & 0.143 & 0.250 & 0.323 \\
\midrule
\multirow{2}{*}{T5} & \multirow{2}{*}{0} & ChEBI & 0.630 & 0.553 & 0.665 & 0.520 & 0.604 & 0.632 \\
& & PCdes & 0.374 & 0.289 & 0.440 & 0.275 & 0.377 & 0.393 \\
\hline
\multirow{2}{*}{MolT5} & \multirow{2}{*}{1 million} & ChEBI & 0.626 & 0.549 & 0.661 & 0.517 & 0.601 & 0.628 \\
& & PCdes & 0.343 & 0.262 & 0.433 & 0.264 & 0.368 & 0.369 \\
\hline
\multirow{2}{*}{\textbf{ChatMol}} & \multirow{2}{*}{6,000} & ChEBI & \textbf{0.647} & \textbf{0.573} & \textbf{0.678} & \textbf{0.538} & \textbf{0.618} & \textbf{0.649} \\
& & PCdes & \textbf{0.396} & \textbf{0.312} & \textbf{0.458} & \textbf{0.295} & \textbf{0.395} & \textbf{0.414}  \\
\bottomrule
\end{tabular}
\caption{Molecule understanding experiment results. \textbf{BL}: BLEU. \textbf{RG}: ROUGE. \textbf{MET}: METEOR. We display the average results under 5 different random seeds. }
\label{tab:M2Tresult}
\end{table*}

\begin{table*}[h]
\centering
\begin{tabular}{l|c|ccccccc}
\toprule
Model & Training Steps & \textbf{EM}↑ & \textbf{hit@3}↑  & \textbf{BL}↑ & \textbf{Leven}↓ & \textbf{RDK}↑ & \textbf{MAC}↑ & \textbf{Morgan}↑ \\
\midrule
ChatGPT & - & 0.009 & 0.009 & 0.405 & 56.27 & 0.283 & 0.483 & 0.184 \\
KV-PLM & 0 & 0.111 & \underline{0.223} & 0.654 & 34.14 & 0.532 & 0.691 & 0.418\\
\midrule
T5 & 0 & 0.052 & 0.078 & 0.600 & 40.71 & 0.513 & 0.672 & 0.419 \\
MolT5 & 1 million & 0.061 & 0.085 & 0.605 & 40.47 & 0.517 & 0.667 & 0.422\\
\textbf{ChatMol} & 6,000 & 0.084	& 0.117	& 0.626	& 38.35 & 0.579 & 0.716	& 0.485 \\
\textbf{ChatMol+} & 6,000 & \textbf{0.140} &	\textbf{0.183} & \textbf{0.712} & \textbf{29.89} & \textbf{0.649} & \textbf{0.764} &	\textbf{0.551} \\
\bottomrule

\end{tabular}
\caption{Molecule generation experiment results. \textbf{EM}: exact matched. \textbf{Leven}: Levenshtein. \textbf{RDK}: RDK FTS. \textbf{MAC}: MACCS FTS. \textbf{Morgan}: Morgan FTS. We display the average results under 5 different random seeds. }
\label{tab:T2Mresult}
\end{table*}

\subsection{Basic Settings}

For molecule understanding, there are few datasets focusing on mutual generation between text and molecule, such as PCdes~\cite{zeng2022deep} and ChEBI-20~\cite{edwardstranslation}. We adopt both as our fine-tuning and evaluation datasets.
There are 26,407 train, 3,301 validation, and 3,300 test pieces of molecule-description pairs in ChEBI-20. Each piece of description has 43.4 words and 3.3 sentences on average. Correspondingly, there are 10,500 train, 1,500 validation, and 3,000 test pieces in PCdes. Descriptions are longer, with 62.1 words and 4.3 sentences on average. 

For molecule generation, we evaluate on our newly-proposed ChEBI-dia conversation dataset. We have 7,361 multi-turn dialogs for training, 1,369 for validation and 1,311 for test. There are altogether 7,626 2-turn dialogs, 4,536 3-turn dialogs, 1,363 4-turn dialogs, 283 5-turn dialogs and 20 even longer dialogs. The sentence characteristics are not much different from the original ChEBI-20 dataset.

We evaluate the following models:

\textbf{ChatGPT}. This refers to the `gpt-3.5-turbo` model in the OpenAI series and is one of the representative LLMs recently. For molecule understanding, we provide 3 randomly-picked examples to tell the exact language style of the specific datasets in the prompt. For molecule generation, we only provide 1 instance to constrain the output format of the model. 

\textbf{KV-PLM}. We adopt the model checkpoint provided in the original paper~\cite{zeng2022deep}, which is a versatile model bridging natural and chemical languages while based on SciBERT~\cite{Beltagy2019SciBERT}, a popular PLM in the biomedical domain. Since KV-PLM is an encoder model, we manage to choose alternative retrieval solutions to compare the capabilities of the generation task as fairly as possible. The model is conducted contrastive learning matching molecules and descriptions. It retrieves the best-matched molecules among train, validation, and test sets for molecule generation, and retrieves sentences from the train set to compose the description paragraphs for molecule understanding.

\textbf{T5}. This is the backbone of our model. We directly fine-tune T5 on the downstream datasets.

\textbf{MolT5}. We adopt the model checkpoint provided in the original paper~\cite{edwardstranslation} which is initialized with T5 checkpoint and then pre-trained by MLM on molecules and texts with 1 million steps (batch size as $256$). 

\textbf{ChatMol}(ours). We initialize ChatMol with T5 and conduct multi-task pre-training. For the molecule understanding version, we adopt MLM, mapping correlation, experimental property and spatial structure tasks and pre-train altogether 6,000 steps (batch size as $256$). For the molecule generation version, we adopt MLM and mapping correlation learning tasks for also altogether 6,000 steps. Notice that the molecule generation version is first tuned on the augmented data produced by the opposite model, and then fine-tuned on ChEBI-dia.

\textbf{ChatMol+}(ours). This is the same model as above, while we allow the plugin use of recognizing entities and querying their SMILES with toolkits and KBs in the molecule generation task, as introduced in Section~\ref{sec:method}. The prompted SMILES can provide mapping knowledge between natural and chemical languages in a targeted manner.

Details for training settings are introduced in Appendix~\ref{sec:setting}.

\begin{table}[ht]\scriptsize
\centering
\begin{tabular}{l}
\toprule
Molecule Understanding \\
\midrule
\texttt{<INPUT>} You are now a chemical assistant. I will provide you with a mole-\\
cule SMILES string, and you have to return a paragraph of proterty descrip-\\
tions for this molecule. For example: \\
    Molecule: CCCC[Si](Cl)(Cl)Cl \\
    Description: it appears as a colorless liquid with a pungent odor. Flash point \\
    126Â°F. Corrosive to metals and skin. Used to make various silicon contain- \\
    ing compounds... \\
    Now the Molecule is: C(C(=O)O)I. What is the Description? \\
\midrule
\texttt{<OUTPUT>} The molecule is an organic compound that belongs to the class of \\
carboxylic acid derivatives. Appears as a colorless to pale yellow liquid with \\
a pungent odor and can be synthesized by treating iodine with acetic anhy-\\
dride. It is known to be corrosive to metals and can cause severe irritation...\\
\toprule
Molecule Generation \\
\midrule
\texttt{<INPUT>} You are now a chemical assistant. I will provide you with a para-\\
graph of property descriptions, and you have to return a SMILES string of a \\
molecule that meet the requirements. For example: \\
Description: It is an acetate ester, a cyclic terpene ketone, a member of \\
furans, a limonoid and a tetracyclic triterpenoid.\\
Molecule: \texttt{CC(=O)O[C@@H]1C[C@@H]2[C@](C=CC(=O)C2(C)C)([C@@H]}\\
\texttt{3[C@@]1(C4=CC[C@H]([C@@]4(CC3)C)[C@@H]5CC(=O)OC5)C)C}\\

Now the Description is: It has a role as a food acidity regulator and a \\
fundamental metabolite. It is an intermediate metabolite in the citric acid \\
cycle. It looks like \texttt{C(C(C(=O)O)C(=O)O)C(=O)O}. What is the Molecule?\\
\midrule
\texttt{<OUTPUT>} The molecule is: \texttt{C(CC(=O)O)C(C(=O)O)C(=O)O}.\\
\bottomrule
\end{tabular}
\caption{Examples of ChatGPT applying to conversational molecular design.}
\label{tab:chat}
\end{table}

\begin{table}[ht]
\centering
\begin{tabular}{l|ccc}
\toprule
Understanding & \textbf{BL-2} & \textbf{BL-4} & \textbf{RG-L}\\
\midrule
Ours & \textbf{0.647} & \textbf{0.573} & \textbf{0.618} \\
\ \ \ \textit{w/o property} & 0.641 & 0.565 & 0.612 \\
\ \ \ \textit{w/o spatial} & 0.640 & 0.565 & 0.613 \\
\ \ \ \textit{w/o mapping} & 0.632 & 0.556 & 0.607 \\
\ \ \ \textit{w/o all} & 0.630 & 0.553 & 0.604 \\
\bottomrule
Generation & \textbf{EM} & \textbf{BL} & \textbf{RDK}\\
\midrule
Ours & \textbf{0.140} & \textbf{0.712} & 0.649 \\
\ \ \ \textit{w/o conversation} & 0.129 & 0.710 & 0.640 \\
\ \ \ \textit{w/o augmentation} & 0.136 & 0.696 & \textbf{0.650}  \\
\ \ \ \textit{w/o mapping} & 0.130 & 0.681 & 0.633 \\
\ \ \ \textit{w/o prompting} & 0.084 & 0.626 & 0.579 \\
\ \ \ \textit{w/o all} & 0.052 & 0.600 & 0.513 \\
\bottomrule
\end{tabular}
\caption{Ablation study results.}
\label{tab:ablation}
\end{table}

\subsection{Result Analysis}

Results for the main experiment are shown in Table~\ref{tab:M2Tresult} and Table~\ref{tab:T2Mresult}. Our model has achieved all the best performances under different settings and metrics. Details for the intermediate-turn results of molecule generation are introduced in Appendix~\ref{sec:inter}.

\textbf{Molecule Understanding Results}. For molecule understanding, our method generates reasonable property descriptions for the given molecules, and achieves comprehensive improvement on both ChEBI-20 and PCdes in all the metrics. As shown in Table~\ref{tab:chat}, the LLM baseline ChatGPT has mastered a part of chemical knowledge and owns a strong natural language generation capability, and therefore can provide property descriptions in a similar style with the given 3 instances. However, it also makes some mistakes (e.g., Iodoacetic acid is not liquid), and the evaluation scores are not comparable with the tuned models, proving the challenge of this task. As for the smaller models, KV-PLM shows no advantages in description generation, and the reason may be that its encoder architecture limits the range of descriptions that the model can obtain. The MLM method-dominated MolT5 adapts worse on PCdes compared with ChEBI-20 the original paper provides, indicating that the over-pre-training may hurt the model generalization capability.

\textbf{Molecule Generation Results}. For molecule generation, our method generates obviously more similar results with ground truth molecules than baselines. The LLM baseline ChatGPT possesses certain molecular knowledge and can always return valid molecules, while the mastered molecules are limited and therefore can not reach a high accuracy under the 1-shot setting. KV-PLM, in contrast, applies the retrieval setting and finds out the best-matched molecules from the given pools, therefore can get higher hit scores, while does not really fit the open scenarios. For the T5-based models, MolT5 shows marginal improvement while not better than ChatMol. Further, augmented by the SMILES obtained from the plugins, our model achieves a jump in performance, proving the importance of bridging different languages and the efficiency of pre-trained models using reliable tools.

\textbf{Ablation study}. We remove our knowledgeable strategies sequentially to show their effectiveness. As shown in Table~\ref{tab:ablation}, \textit{w/o property}, \textit{w/o spatial}, and \textit{w/o mapping} refer to the removal of experimental property prediction, spatial structure prediction, and molecule mapping correlation in the multi-task pre-training; \textit{w/o conversation} refers to the removal of molecules in the conversation history; \textit{w/o augmentation} refers to removing the dual augmentation training; \textit{w/o prompting} refers to removing the SMILES prompts obtained from plugins.

As we can see, all the non-full versions of ChatMol perform worse on both tasks, proving the effectiveness of our method. Especially the mapping correlation training and the SMILES prompting show significantly lower scores when being removed, showing that the capability of bridging versatile subdomains is essential to accomplish the conversational molecular design. The \textit{w/o conversation} version proves that the iterative modification form is more reasonable than directly providing a whole paragraph of text requirements.

\begin{figure*}[ht]
    \centering
    \includegraphics[width=1\linewidth]{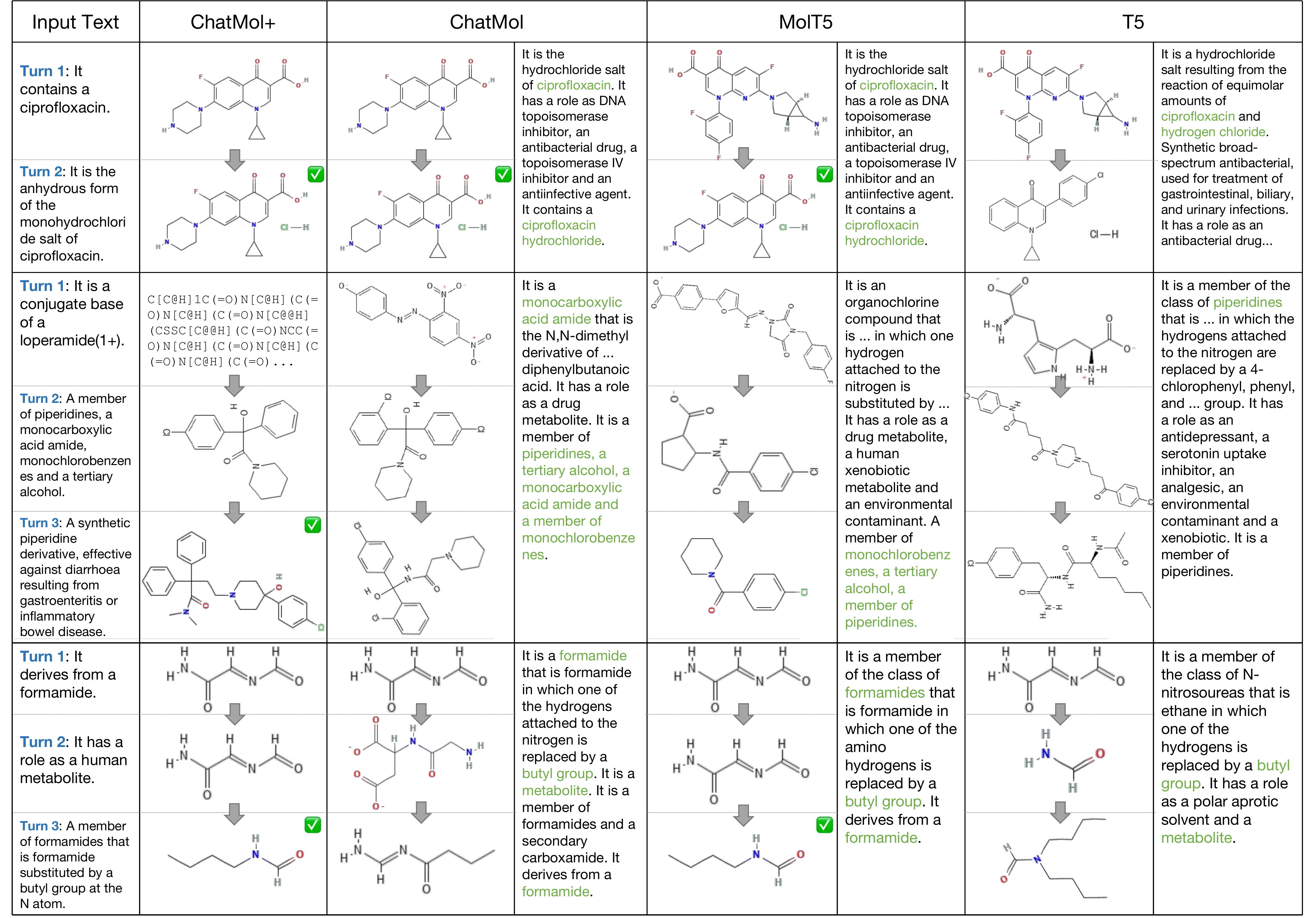}
    \caption{Case study for conversational molecular design. The correct molecules and properties are marked in green. }
    \label{fig:case}
\end{figure*}

\subsection{Case Study}

We provide several generation cases from the ChEBI-dia test set in Figure~\ref{fig:case}. Both the intermediate results for the multi-turn input text descriptions and the generated property descriptions given the ground-truth molecules are shown. 

As we can see, for the easy case (line 1) that contains clear structural descriptions, all models provide high-quality molecule outputs from the first turn, and modify the molecule with the hydrochloride group according to the supplemented second turn information. For the comparably difficult case (line 2) that provides more general and vague descriptions, baseline models fail to generate correct molecules and precise descriptions, while ChatMol+ eventually succeeds even when it makes a mistake in the first turn (which is not a valid molecule). ChatMol also grasps some key required substructures (e.g., piperidine, tertiary alcohol, monochlorobenzene) and generates a quite similar molecule, and meanwhile provides much more concise descriptions compared with baseline models. Nevertheless, the models trained in a targeted manner (MolT5 and our models) have overall better performances than T5 in many cases, as shown in line 3.

Imagine the future where we can apply the technology of conversational molecular design for drug discovery, describing our command for structures and functions and obtaining possible molecules from the model outputs. Besides, readable descriptions can be generated when chemical researchers or learners input any unfamiliar molecules just as we do in the case study. Currently, the model and the data scales are both small, and the modifications and descriptions are also not so granular. It is possible to get stronger conversational molecular design systems that apply larger models with more fine-grained training data, and have powerful knowledge memory to do more reliable predictions.

\section{Related Work}

\subsection{Knowledgeable Models}
Though the deep learning models have the ability to absorb commonsense knowledge from large-scale training data in an unsupervised manner, directly injecting knowledge into models still shows its effectiveness. Specific to the biochemical domain, there are plenty of PLMs that are transferred from the general domain by simply replacing their corpus~\cite{wang2021pre}, while knowledgeable training and inferencing methods still exist. There are two mainstream approaches: (1) Explicitly encode the knowledge to assist the models. Extra modules are usually targeting designed for the downstream task. For instance, related tags can be acquired from knowledge graphs (KGs) and separately encoded to get knowledge fusion attention for the chemical-protein interaction extraction~\cite{sun2020chemical}. KGs information can also be incorporated via a hierarchical graph representation to modify the PLM contextualized embeddings for biomedical entity/relation/event extraction~\cite{huang2020biomedical,lai2021joint}. (2) Implicitly teach the knowledge with special pre-training tasks. The knowledge items can help prepare data for multi-task training with methods such as distant supervision. Based on KG items, a binary classification task is added to predict whether a triplet exists or not~\cite{hao2020enhancing}, and entity detection and linking tasks are conducted for similar purposes~\cite{yuan2021improving}. 

When it comes to SMILES strings processing, supplementary knowledge especially spatial knowledge is also emphasized since SMILES cannot fully represent the topological information of molecules~\cite{zhang2021graph}. Graph representations~\cite{elton2019deep} are popular with molecular deep learning research since they do not have to consider rotation, permutation and other issues additionally compared with other methods such as atom coordinates. 
Some graph-based methods also incorporate 3D spatial information~\cite{danel2020spatial}, using the universal force field(UFF) to generate the conformations. GEM~\cite{fang2022geometry} explores the effect of using different force field constraints(e.g. MMFF~\cite{halgren1996merck} and DFT~\cite{parr1983density}) on the quality of the generated coordinates.

\subsection{Models Bridging Versatile Subdomains}
There exist countless cross-modal methods, especially in the image-text field. Pre-trained models such as CLIP~\cite{radford2021learning} and Unimo~\cite{li2021unimo} have brought the fusion of the two modals to a new level. With similar ideas, we can also bridge data from various subdomains that have natural gaps. 

To process different forms of biochemical data, one solution is to convert them into the same modal and form. Goh et al.~\cite{goh2018using} unify the molecular graph data and SMILES data to molecular structure images as the model input. KV-PLM~\cite{zeng2022deep} is the first research trying to fuse SMILES strings and natural language descriptions together to improve the knowledge reserve and chemical comprehension of the model. MolT5~\cite{edwardstranslation} further develops the idea to conduct translation between molecules and text descriptions by a generative model.
Su et, al.~\cite{su2022molecular} propose a cross-modal graph-based model linking molecular and natural language, while it ignores the flexibility of natural language descriptions. Overall, bridging chemical and natural language is an innovative scenario that has not been fully explored.

\section{Conclusion}

In this paper, we propose conversational molecular design, which is a novel interactive paradigm adopting natural language for describing and editing target molecules. Two specific tasks, molecule understanding and molecule generation, are explored, and the dataset ChEBI-dia is made. We present ChatMol, a knowledgeable generative model bridging chemical and natural language descriptions for molecules. By injecting molecular knowledge and bridging different languages, our model is further enhanced and proven to be more effective than existing methods with significantly less training cost. This technique promises to drive a new paradigm in AI-assisted molecular design. 

\section*{Code and Data Availability}
Codes and data are provided in \url{https://github.com/Ellenzzn/ChatMol/tree/main}. 

\section*{Limitations}
Though proved to be effective and efficient, our method is temporarily not validated on a larger scale of model parameters and training data. In fact, the large version of our baseline (MolT5) is reported to be much more powerful than the small and base versions, and this may indicate that various types of knowledge that we have discussed can be automatically learned and mastered once the computing and data resources are plenty. 

Better evaluation methods are also urgently needed. As we have observed, some descriptions are correct while not recorded in the ground truth texts, and conversely, a paragraph of descriptions might not only correspond to a single molecule. Therefore, it is hardly possible to thoroughly and reasonably evaluate model performance currently.

\section*{Ethics Statement}
The authors declare that they have no conflict of interest. This article does not contain any studies with animals or human beings performed by any of the authors. 

\bibliography{anthology,custom}
\bibliographystyle{acl_natbib}

\appendix

\section{Pre-training Corpus}
\label{sec:dataset}
For the pre-training period, we obtain the text corpus from S2orc~\cite{lo2020s2orc}, which contains over 12.7 million academic papers. We download SMILES strings from the PubChem database~\cite{kim2016pubchem}, which contains billions of substance records. The data between different tasks do not overlap.

For the SMILES corpus, we conduct three different tasks. (1)~\textit{SMILES MLM}: We sample SMILES strings for 300k molecules to conduct the basic MLM task. (2)~\textit{Molecular property prediction}: We collect 98k items of physical and chemical properties from wet-lab experiments of 31k molecules from PubChem. There are altogether 15 types of properties including Solubility, Color/Form, Boiling Point, Flash Point, Density, Vapor Density, Decomposition, Corrosivity, Melting Point, LogP, Vapor Pressure, Stability/Shelf Life, Odor, Taste, and pH. Notice that we also remain the SMILES-name pairs of these molecules to conduct mapping correlation learning. (3)~\textit{Spatial structure learning}: We retain the other 300k molecules from PubChem. 15\% of the atoms in each molecule are randomly marked and required to recognize their connected atoms, aromaticity and ring formation information.

For the text corpus, we conduct two types of tasks. (1)~\textit{Text MLM}: We select 400k literature abstracts in the chemistry and biomedicine field for the basic MLM task. (2)~\textit{Mapping correlation learning}: There are two methods for automatically obtaining parallel data between text and molecules. The first one is to directly match the name-SMILES pairs, and we sample 200k molecules for it. The second one is to roughly annotate the corresponding entity SMILES in the corpus. Specifically, we adopt SciSpacy~\cite{neumann2019scispacy} to find out the chemical entities, which are then matched with the list of substances and synonyms from PubChem. In this way, we get over 2.4 million mentions of 4.7k most frequently-appeared molecules linked with their SMILES strings in 760k literature abstracts. 

For the dual augmentation, we adopt the 14.7k molecules from ChEBI-20 train set that are not contained in ChEBI-dia as the augmented pool, while do not use their text descriptions to avoid introducing extra information.

\section{Intermediate turn analysis}
\label{sec:inter}

We provide the intermediate prediction results of the multi-turn molecule generation task in Table~\ref{tab:intermediate}. Since the intermediate answers are annotated automatically, their accuracy cannot be guaranteed and should be used as a reference only. From the results we can conclude that: (1) We use the MolT5 initial checkpoint to annotate intermediate answers, and naturally the fine-tuned MolT5 generates similar molecules and achieves a quite satisfying performance. Though ChatMol gets a lower score, the plugin version ChatMol+ does even better than MolT5, showing the great assistance that the chemical-natural language mapping tool (chemical entity recognition toolkits and SMILES KBs) can provide; (2) Early turns have high accuracy, indicating that it is easy for the models to find molecules that meet the requirements for the general descriptions, while it is more difficult to complete the modification for the supplementary fine structure descriptions.

\begin{table}
\centering
\begin{tabular}{l|c|ccc}
\toprule
Model & Turn & \textbf{EM} & \textbf{BL} & \textbf{RDK}\\
\midrule
\multirow{4}{*}{ChatMol+} & 1 & \textbf{0.402} & \textbf{0.716} & \textbf{0.672} \\
 & 2 & \textbf{0.211} & \textbf{0.691} & \textbf{0.592} \\
 & 3 & \textbf{0.190} & \textbf{0.716} & \textbf{0.614} \\
 & 4 & 0.125 & 0.716 & 0.626 \\
\midrule
 \multirow{4}{*}{ChatMol} & 1 & 0.393 & 0.568 & 0.638 \\
 & 2 & 0.192 & 0.532 & 0.573 \\
 & 3 & 0.092 & 0.522 & 0.578 \\
 & 4 & 0.176 & 0.587 & \textbf{0.636} \\
\midrule
 \multirow{4}{*}{MolT5} & 1 & 0.384 & 0.689 & 0.635 \\
 & 2 & 0.194 & 0.650 & 0.559 \\
 & 3 & 0.135 & 0.657 & 0.536 \\
 & 4 & \textbf{0.206} & \textbf{0.717} & 0.598 \\
\midrule
 \multirow{4}{*}{T5} & 1 & 0.367 & 0.676 & 0.630 \\
 & 2 & 0.162 & 0.630 & 0.540 \\
 & 3 & 0.130 & 0.669 & 0.539 \\
 & 4 & 0.118 & 0.678 & 0.503 \\
\bottomrule
\end{tabular}
\caption{Intermediate results for molecule generation.}
\label{tab:intermediate}
\end{table}

\section{Training Setting Supplementary}
\label{sec:setting}

We initialize ChatMol with the pre-trained T5-base (with the hidden size of $768$). Other hyperparameter details can be found in the original paper~\cite{raffel2020exploring}. Molecule understanding and generation correspond to two different checkpoints. We implement our method in the PyTorch~\cite{paszke2019pytorch} framework, and adopt the Huggingface Transformers~\cite{wolf2019huggingface}. We take the AdamW optimizer~\cite{loshchilov2019decoupled} which is suitable for most of the PTMs in the T5 backbone.

For the multi-task pre-training, we add different task prefixes (e.g. ``Predict Solubility:'') to the input, and randomly mix up the training data for MLM, SMILES-text parallel generation, property and spatial information prediction. In this period, we set the learning rate as $5e-4$ and batch size as $256$.

For the downstream molecule understanding fine-tuning, we set the maximum epoch number as $50$, and the early stop epoch number as $5$. The learning rate and batch size is searched in $\left\{1e-3, 5e-4, 2e-4\right\}$ and $\left\{8, 16, 32\right\}$, and eventually decided as $5e-4$ and $16$. Notice that our hyper-parameter settings are different from the MolT5 paper~\cite{edwardstranslation}, therefore the results are higher than original reported. For molecule generation, the learning rate and batch size is searched in $\left\{1e-3, 5e-4, 1e-4, 5e-5\right\}$ and $\left\{16, 32, 64\right\}$, and eventually decided as $5e-4$ and $32$.  

We train our models on a NVIDIA A100 SXM4 40 GB GPU. Evaluation metrics such as BLEU and ROUGE are achieved with the assistance of NLTK toolkit~\cite{loper2002nltk}.

\section{Licenses and Data Usage Policy}
PCdes, ChEBI-20 and S2orc are released under the CC BY-SA 4.0 license. MoleculeNet are released under the MIT license. All the datasets are used in a way consistent with their intended use. We observe the data samples and do not find any offensive content or identifiers in these datasets.

\end{document}